\documentclass[10pt,twocolumn,letterpaper]{article}

  \usepackage[pagenumbers]{cvpr} 

\usepackage{booktabs}
\usepackage{graphicx}
\usepackage{float}

\definecolor{cvprblue}{rgb}{0.21,0.49,0.74}
\usepackage[pagebackref,breaklinks,colorlinks,allcolors=cvprblue]{hyperref}

\def\paperID{*****} 
\def\confName{CVPR}
\def\confYear{2026}

\title{ Parameter Efficient Fine-tuning for Domain-specific Gastrointestinal Disease Recognition}

\author{Sanjaya Poudel\textsuperscript{1,*}  \quad 
Nikita Kunwor\textsuperscript{2} \quad 
Raj Simkhada\textsuperscript{2} \quad
Mustafa Munir\textsuperscript{3} \\
Manish Dhakal\textsuperscript{4} \quad
Khem Poudel\textsuperscript{5,\thanks{Correspondence to: Sanjaya Poudel $<$szp0223@auburn.edu$>$; Khem Poudel $<$khem.poudel@mtsu.edu$>$}} \quad
\\
\normalsize
\textsuperscript{1}Auburn University \quad
\textsuperscript{2}Tribhuvan University \quad
\textsuperscript{3}The University of Texas at Austin \quad 
\textsuperscript{4}Georgia State University \\
\normalsize
\textsuperscript{5}Middle Tennessee State University \\
}

\begin{document}
\maketitle
\begin{abstract}

Despite recent advancements in the field of medical image analysis with the use of pretrained foundation models, the issue of distribution shifts between cross-source images largely remains adamant.
To circumvent that issue, investigators generally train a separate model for each source.
However, this method becomes expensive when we fully fine-tune pretrained large models for a single dataset, as we must store multiple copies of those models.
Thus, in this work, we propose using a low-rank adaptation (LoRA) module for fine-tuning downstream classification tasks.
LoRAs learn lightweight task-specific low-rank matrices that perturb pretrained weights to optimize those downstream tasks.
For gastrointestinal tract diseases, they exhibit significantly better results than end-to-end finetuning with improved parameter efficiency. Code is available at: \url{https://github.com/sanjay931/peft-gi-recognition}.
\end{abstract}    
\section{Introduction}
\label{sec:intro}

Medical image analysis has become a vital area of research and is essential for diagnosing and predicting many medical conditions. While these images make up the majority of medical data today, the development of robust computer-aided systems frequently suffers from failed experimentation due to dataset distribution shifts~\cite{yoon2024domain,guan2021domain,guo2024impact,yang2024limits,zhang2020generalizing}. When models trained on one dataset are applied to a different source dataset for the same task, variations in image quality, brightness, and other domain-specific features can cause a severe degradation in diagnostic performance.

To solve this issue, it often becomes necessary to train separate, specialized models tailored to the distinct characteristics of separate data sources. As indicated in \cref{tab:cross_source_testing}, which details our cross-source testing, a model's inability to seamlessly generalize across different clinical environments highlights the necessity of source-specific adaptation to maintain high classification accuracy.

\begin{table}[h]
    \centering
    \resizebox{0.9\linewidth}{!}{ 
        \begin{tabular}{l|cc}
        \hline
        \textbf{Train \textbackslash\ Test} & \textbf{GastroVision } & \textbf{Kvasir-Capsule } \\
        \hline
        \textbf{GastroVision~\cite{jha2023gastrovisionmulticlassendoscopyimage}} & 86.31 & 8.04 \\
        \textbf{Kvasir-Capsule~\cite{Smedsrud2021}} & 10.12 & 99.05 \\
        \hline
        \end{tabular}
    }
    \caption{\textbf{Cross-domain evaluation of ConvNeXt-V2-Tiny~\cite{woo2023convnextv2}:} Cross-domain (Non-diagonal) accuracy demonstrates performance drops due to distribution shifts.}
    \label{tab:cross_source_testing}
\end{table}

While training models specifically for each target source resolves the issue of distribution shifts, end-to-end training remains a resource-exhaustive effort. To conserve valuable hardware resources and time, transfer learning (TL)~\cite{tang2022self} is widely implemented, allowing models to leverage knowledge gained from a large pretrained task to improve performance on a new, specialized target task. However, fine-tuning large, advanced convolutional neural networks (CNNs) like ConvNeXt-V2~\cite{woo2023convnextv2} can require a lot of computational power and memory. Due to the immense resource demands of modern models, the traditional practice of updating the entire model, known as end-to-end fine-tuning, is the most resource-intensive TL approach.

To address this, we propose to use an existing highly efficient framework for parameter-efficient fine-tuning (PEFT) that incorporates low-rank adaptation (LoRA)~\cite{hu2022lora} to adapt the ConvNeXt-V2 model for domain-specific (or source-specific) image classification tasks. 
LoRA helps fine-tune large pre-trained models effectively while dramatically reducing the number of parameters that need to be trained. 
This approach uses LoRA to add only a small number of trainable parameters, leaving most of the pre-trained model weights frozen. This helps reduce memory usage and speeds up the training process, making it easier to fine-tune large models.

Our contributions are as follows:
\begin{itemize}
    \item We propose to use PEFT module, specifically LoRA, with ConvNeXt-V2 backbone for domain-specific image classification, effectively adapting the model while keeping the pretrained weights frozen.
    \item We design an end-to-end training pipeline featuring data augmentation and optimized cross-entropy feedback.
\end{itemize}
\section{Related Work}
\label{sec:RelatedWork}

Our work relates to Convolutional Neural Networks (CNNs), the Transformer architecture, Parameter-Efficient Fine-Tuning (PEFT) methods like LoRA, and the classification of gastrointestinal (GI) tract diseases and findings. Here, we thoroughly review the literature that bears significant relevance to these areas.
\subsection{Gastrointestinal disease}

The availability of public datasets is crucial for research in computer-aided diagnosis (CAD) systems for GI diseases. The Kvasir-Capsule dataset represents a significant resource comprising thousands of still images extracted from video capsule endoscopy, covering a wide range of gastrointestinal conditions. Building on this foundation, GastroVision was recently presented as a multi-center open-access GI endoscopy dataset, containing 8,000 images categorized across 27 classes of landmarks, abnormalities, and findings \cite{jha2023gastrovisionmulticlassendoscopyimage}.

Early computational work leveraged standard Convolutional Neural Networks (CNNs) for classification; for instance, fine-tuned residual models (ResNet50) and dense models (DenseNet121) achieved test accuracies around 96.96\% on the Kvasir-Capsule dataset \cite{alibouch2025two}. More advanced CNN structures have been explored, such as the presentation of Denoising Capsule Networks (Dn-CapsNets) by Afriyie et al., which achieved an improved accuracy of 94.16\% on the Kvasir-v2 dataset~\cite{article}. Recent models have begun integrating different architectures; for example, a hybrid approach combining DenseNet201 (as a CNN feature extractor) with a Swin Transformer branch was proposed to utilize both local and global feature understanding for classification on the GastroVision and Kvasir-Capsule datasets \cite{subedi2024classificationendoscopyvideocapsule}. Furthermore, Thambawita et al. proposed a system for classifying disease in the gastrointestinal tract based on global features and deep neural networks \cite{thambawita2018medicotask2018diseasedetection}.

\subsection{Parameter-Efficient Fine-Tuning and ConvNeXt}

Traditional CNNs, while effective, utilize pooling layers that prioritize translational invariance, which can lead to the loss of significant information. In contrast, the Transformer architecture has gained significant attention in computer vision due to its proficiency in modeling long range dependencies~\cite{huang2024incontextloradiffusiontransformers}.
However, adapting large pre-trained vision models (whether pure Transformers or modern CNN variants like ConvNeXt V2-B, which is used in recent gastroscopy analysis) to specialized medical datasets often faces challenges related to computational cost and data scarcity \cite{kim2022transfer,gao2023lesscompactconvolutionaltransformers}. This difficulty led to the rise of parameter-efficient fine-tuning (PEFT) methods .
Low-Rank Adaptation (LoRA), a leading PEFT technique initially proposed by Hu et al., addresses this by freezing the majority of the large pre-trained model weights and injecting small, trainable low-rank decomposition matrices (adapters)~\cite{Zhu2025PDoRA}. LoRA has proven highly effective in specialized medical classification tasks, demonstrating that foundation models can be adeptly tailored: for example, adapting the DINOv2 foundation model with LoRA achieved a high accuracy of 97.75\% on the Kvasir-Capsule dataset \cite{zhang2024learningadaptfoundationmodel}. Crucially for our work, Convolutional Low-Rank Adaptation (CoLoRA) successfully extends LoRA’s efficiency benefits to CNN backbones like VGG16 and ResNet50 by decomposing kernel updates, offering a stable and resource-efficient alternative to full fine-tuning for medical image classification~\cite{rivera2025coloraefficientfinetuningconvolutional}. This demonstrates the feasibility and effectiveness of combining a powerful convolutional architecture (like ConvNeXt) with LoRA for specialized, multi-level classification tasks. Similarly, Jiamin Hu et al.~\cite{10.1007/978-981-96-3863-5_15} proposed LoRA-MedSAM, where the pre-trained MedSAM weights are kept frozen, and only the LoRA parameters are updated.

\section{Methodology}
\label{sec:Methodology}
This section presents a concise description of Parameter-Efficient Fine-Tuning incorporating Low-Rank Adaptation (LoRA)~\cite{hu2022lora} to adapt the ConvNeXt-V2 model~\cite{woo2023convnextv2} for medical image classification. Specifically, \Cref{sec:convnext_base_arch} details the ConvNeXtV2-Base backbone architecture, including its hierarchical stage design and Global Response Normalization MLP blocks. \Cref{sec:lora_integration} describes the integration of LoRA adapters into the linear projection layers of the backbone to enable efficient fine-tuning with a minimal number of trainable parameters.
\begin{figure}[H]
  \centering
   \includegraphics[width=0.95\linewidth]{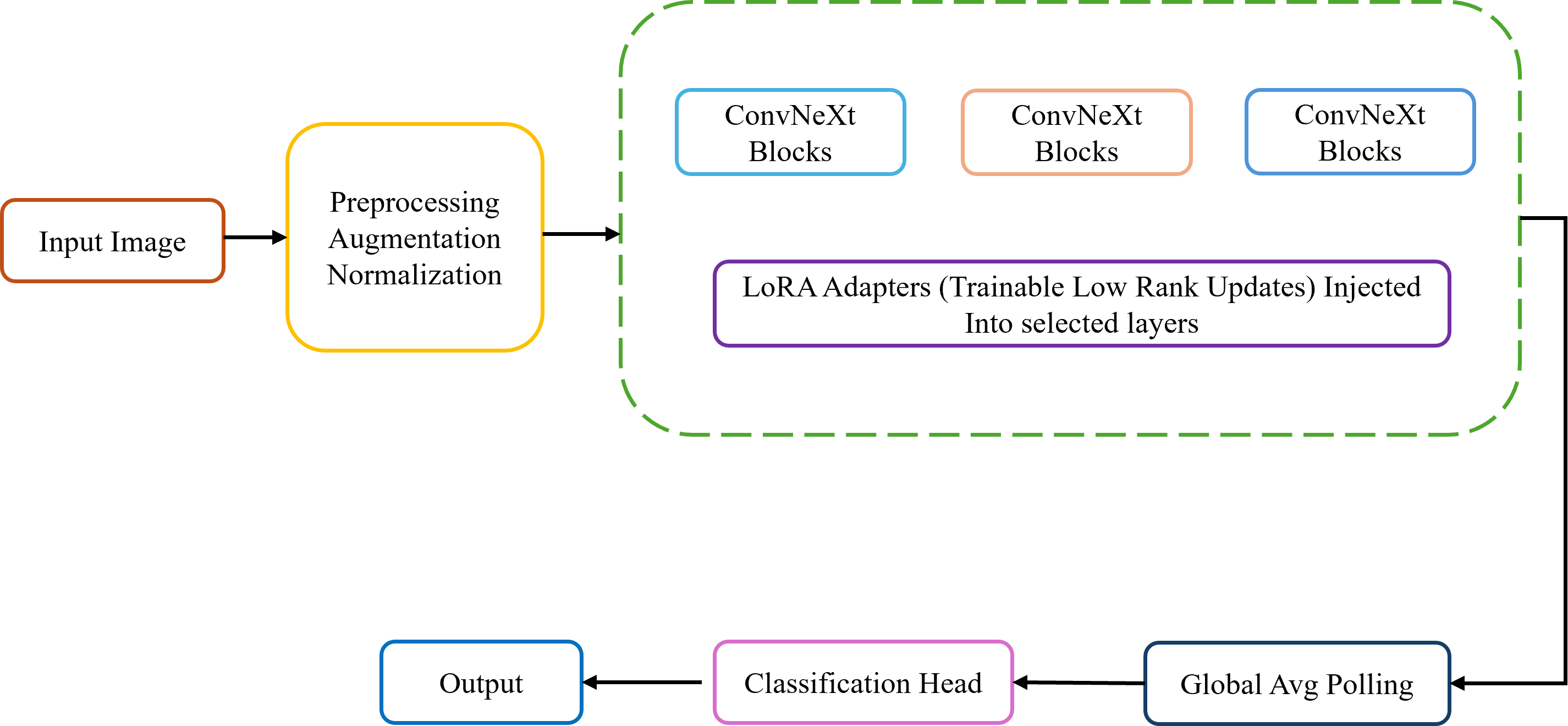}
   \caption{System Architecture of PEFT using ConvNeXt-Base.}
   \label{fig:onecol}
\end{figure}

\Cref{fig:onecol} illustrates the system architecture for parameter-efficient fine-tuning (PEFT) using ConvNeXtV2 as the backbone model. The architecture incorporates Low-Rank Adaptation (LoRA) to optimize the model by introducing small, trainable low-rank matrices into the ConvNeXtV2 layers, significantly reducing the number of parameters that need to be trained, while leaving most of the pretrained weights frozen for efficient adaptation to domain-specific tasks.

\subsection{ConvNeXt-Base Architecture}
\label{sec:convnext_base_arch}
   We use ConvNeXtV2-Base as the backbone network. The architecture is composed of four hierarchical stages with channel dimensions \([128, 256, 512, 1024]\) and block distribution \([3, 3, 27, 3]\), yielding approximately \(89\)M parameters. The model starts with a patch embedding stem implemented as a \(4 \times 4\) convolution with stride \(4\), followed by progressive downsampling between stages using LayerNorm and \(2 \times 2\) convolutions.

Each stage consists of repeated ConvNeXt blocks, where spatial feature extraction is performed by a \(7 \times 7\) depthwise convolution, followed by LayerNorm and an MLP-style transformation module. In our implementation, this transformation module is a \textit{Global Response Normalization MLP}, containing two linear projection layers (\texttt{fc1} and \texttt{fc2}), a GELU activation~\cite{hendrycks2016gaussian}, and a GRN layer. Residual connections are preserved throughout the network. After the final stage, global average pooling and a normalized classification head produce the final class predictions.

\subsection{LoRA Integration}
\label{sec:lora_integration}
To efficiently adapt the pretrained ConvNeXtV2 backbone to our task, we employ \textbf{Low-Rank Adaptation (LoRA)}~\cite{hu2022lora}, a parameter-efficient fine-tuning technique that significantly reduces the number of trainable parameters. Instead of updating the full set of model weights during training, LoRA introduces small trainable low-rank matrices into selected layers while keeping the original pretrained weights frozen.

In the ConvNeXtV2 architecture, each block contains a lightweight MLP transformation module consisting of two linear projection layers, denoted as \texttt{fc1} and \texttt{fc2}. These layers perform channel-wise feature transformations and therefore represent effective points for parameter-efficient adaptation. In our implementation, LoRA adapters are inserted into these projection layers across all ConvNeXt blocks.

Formally, given a weight matrix \(W \in \mathbb{R}^{d \times k}\), LoRA reparameterizes the weight update as

\[
W' = W + \Delta W
\]

where

\[
\Delta W = BA
\]

with

\[
A \in \mathbb{R}^{r \times k}, \quad B \in \mathbb{R}^{d \times r}
\]

and \(r \ll \min(d,k)\). During fine-tuning, the original weight \(W\) remains frozen while only the low-rank matrices \(A\) and \(B\) are optimized. This formulation allows the model to adapt to new tasks while maintaining the pretrained representation.

In our experiments, we set the LoRA rank to \(r = 16\) and the scaling factor to \(\alpha = 32\). Additionally, a LoRA dropout of \(0.1\) is applied to improve generalization and reduce overfitting. With this configuration, only a small fraction of the model parameters are updated during training, enabling efficient fine-tuning while preserving the majority of pretrained weights.

\begin{figure}[H]
  \centering
   \includegraphics[width=.9\linewidth]{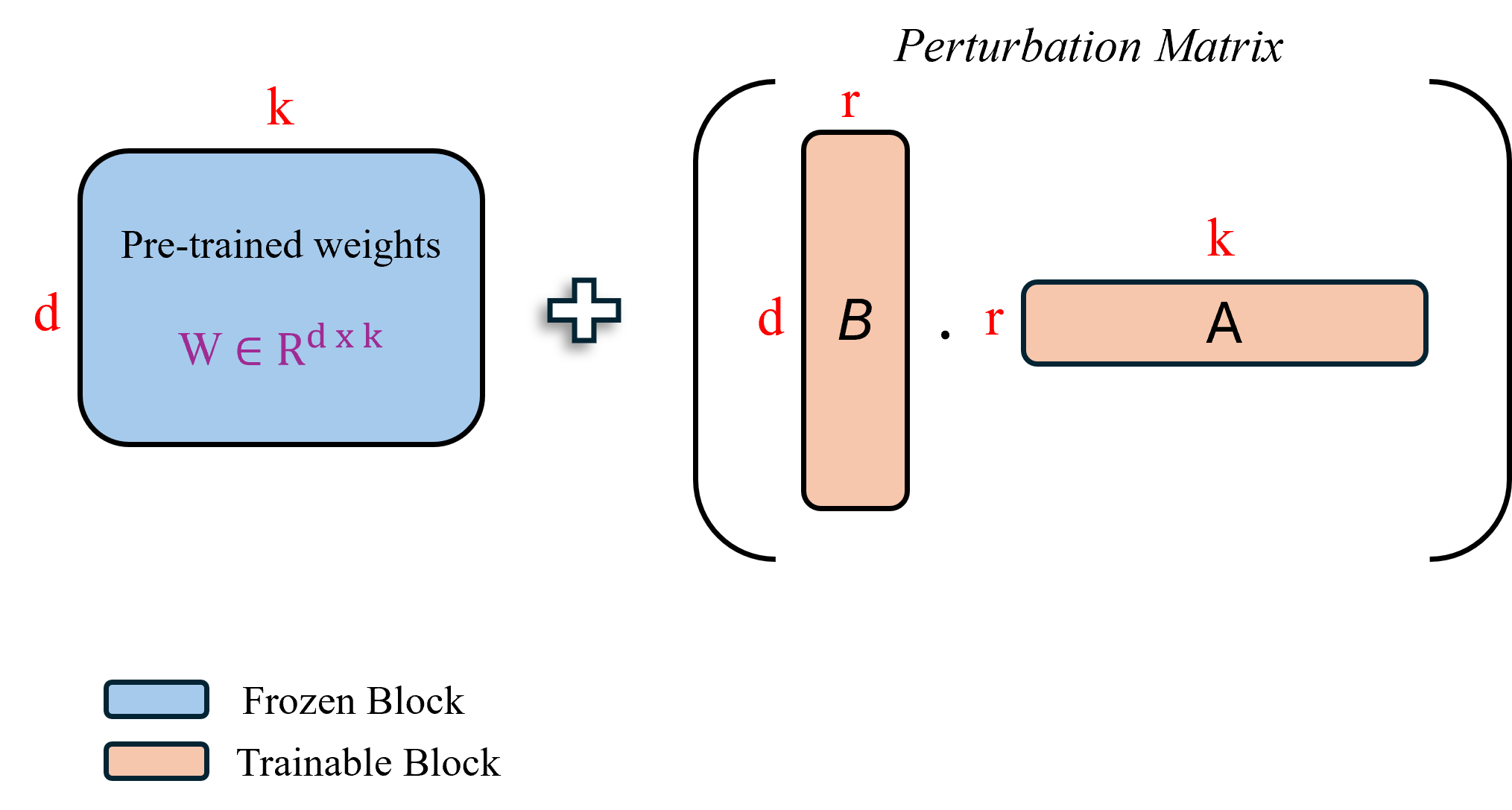}
   \caption{LoRA-based weight updates, where we only train $A$ and $B$ meanwhile keeping the whole pretrained model frozen.}
   \label{fig:lora}
\end{figure}

\Cref{fig:lora} demonstrates the LoRA-based weight update process in the ConvNeXtV2 model. The figure illustrates that only the low-rank matrices (A and B) are trained during fine-tuning, whereas the remaining pretrained model weights remain frozen. This efficient adaptation reduces computational burden and memory usage, enabling fine-tuning of specialized tasks with minimal parameter updates.

\section{Experiment}
\label{sec:Experiment}
\subsection{Datasets}

We evaluate our method on two publicly available gastrointestinal image classification datasets: GastroVision and Kvasir-Capsule.

\subsubsection{GastroVision}

GastroVision~\cite{jha2023gastrovisionmulticlassendoscopyimage} is a multi-class gastrointestinal endoscopy image dataset designed for computer-aided GI disease detection. The dataset contains 8,000 images spanning 27 labeled classes, including anatomical landmarks, pathological findings, therapeutic intervention cases, and normal observations from the gastrointestinal tract. The images were collected from two medical centers, Bærum Hospital in Norway and Karolinska University Hospital in Sweden, and were annotated and verified by experienced gastrointestinal endoscopists. Most images were acquired using white-light imaging, while a smaller portion was obtained using narrow-band imaging.

In this work, we use a subset of 20 classes from the GastroVision dataset, including normal mucosa and vascular pattern in the large bowel, dyed-resection margins, duodenal bulb, normal stomach, Barrett's esophagus, small bowel terminal ileum, normal esophagus, accessory tools, cecum, resected polyps, retroflex rectum, colorectal cancer, gastric polyps, colon polyps, pylorus, dyed-lifted polyps, gastroesophageal junction normal Z-line, blood in lumen, esophagitis, and ileocecal valve. The dataset was divided into training, validation, and test sets using an 80\% / 10\% / 10\% split. Similar to many medical imaging datasets, the class distribution is imbalanced, with several categories containing relatively fewer samples than others.

\subsubsection{Kvasir-Capsule}

Kvasir-Capsule~\cite{Smedsrud2021} is a large video capsule endoscopy (VCE) dataset collected from clinical examinations at Bærum Hospital in Norway. The full dataset contains 117 videos with a total of 4,741,504 frames. Among these, 47,238 frames are labeled and medically verified with bounding-box annotations across 14 classes, while the remaining frames are unlabeled. The dataset focuses on small-bowel capsule endoscopy and includes both anatomical landmarks and luminal/pathological findings.

In this work, we use a subset of 11 classes from the labeled portion of Kvasir-Capsule, namely erythema, ulcer, reduced mucosal view, pylorus, normal clean mucosa, lymphangiectasia, blood-fresh, ileocecal valve, erosion, foreign body, and angiectasia. To reduce information leakage caused by highly correlated neighboring frames, the data were split at the video level into training, validation, and test sets. Similar to many medical imaging datasets, the class distribution is imbalanced, with some findings containing substantially fewer labeled samples than others.

\subsection{Implementation Details}

All experiments were implemented using the PyTorch deep learning framework and trained on an A100 NVIDIA GPU. We adopt ConvNeXtV2-Base as the backbone network, initialized with ImageNet pretrained weights. To enable parameter-efficient fine-tuning, LoRA adapters are inserted into the linear projection layers of the ConvNeXt blocks. The LoRA configuration uses rank $r=16$, scaling factor $\alpha=32$, and a dropout rate of $0.1$.

Two separate models were trained corresponding to the two datasets used in this study: GastroVision and Kvasir-Capsule. For both datasets, input images were resized to $224 \times 224$ pixels before being fed into the network. Standard data augmentation techniques including random horizontal flipping and small rotations were applied during training to improve generalization.

Training was performed using the AdamW optimizer with an initial learning rate of $1\times10^{-4}$. Models were trained for a maximum of 30 epochs with a batch size of 32. To prevent overfitting and unnecessary training, early stopping with a patience of 5 epochs was applied based on validation performance. The model achieving the best validation accuracy was selected for final evaluation.

\subsection{Evaluation Metrics}

To evaluate model performance, we report Accuracy, Precision, Recall, F1-score, and the Matthews Correlation Coefficient (MCC), which are commonly used metrics for multi-class classification tasks.

Accuracy measures the proportion of correctly classified samples among all predictions:

\begin{equation}
Accuracy = \frac{TP + TN}{TP + TN + FP + FN}
\end{equation}

Precision represents the proportion of correctly predicted positive samples among all predicted positives:

\begin{equation}
Precision = \frac{TP}{TP + FP}
\end{equation}

Recall (also called sensitivity) measures the proportion of correctly predicted positive samples among all actual positives:

\begin{equation}
Recall = \frac{TP}{TP + FN}
\end{equation}

The F1-score is the harmonic mean of Precision and Recall:

\begin{equation}
F1 = 2 \times \frac{Precision \times Recall}{Precision + Recall}
\end{equation}

Finally, we report the Matthews Correlation Coefficient (MCC), which provides a balanced evaluation even when class distributions are imbalanced:

\begin{equation}
\resizebox{0.45\textwidth}{!}{%
$MCC = \frac{TP \times TN - FP \times FN}{\sqrt{(TP+FP)(TP+FN)(TN+FP)(TN+FN)}}$
}
\end{equation}

These metrics provide complementary perspectives on classification performance and are reported for both the GastroVision and Kvasir-Capsule datasets.

\subsection{Results}
\label{sec:result}

\begin{table*}[t]
\centering
\caption{Quantitative comparison of different models on the GastroVision and Kvasir-Capsule datasets.}
\label{tab:results}
\resizebox{0.95\textwidth}{!}{%
\begin{tabular}{lrcccccccccc}
\toprule
& & \multicolumn{5}{c}{\textbf{GastroVision}} & \multicolumn{5}{c}{\textbf{Kvasir-Capsule}} \\
\cmidrule(lr){3-7} \cmidrule(lr){8-12}
\textbf{Method} & \textbf{\#Params.} & Prec. & Rec. & F1 & Acc. & MCC & Prec. & Rec. & F1 & Acc. & MCC \\
\midrule
MobileNetV2 ~\cite{sandler2018mobilenetv2} 
& 3.5M & 74.74 & 76.67 & 74.53 & 76.67 & 73.88
& 95.34 & 95.49 & 95.22 & 95.49 & 90.51 \\

Inception V3 ~\cite{szegedy2015rethinkinginceptionarchitecturecomputer}
& 25.4M & 79.87 & 80.44 & 79.31 & 80.44 & 78.17
& 97.27 & 97.41 & 97.32 & 97.41 & 94.57 \\

DenseNet121 ~\cite{huang2017densely}
& 7.5M & 82.57 & 83.17 & 82.12 & 83.17 & 81.14
& 97.42 & 97.59 & 97.47 & 97.59 & 94.93 \\

DenseNet201 ~\cite{huang2017densely} 
& 19.1M & 83.14 & 82.91 & 81.82 & 82.91 & 80.92
& 98.62 & 98.60 & 98.57 & 98.60 & 97.07 \\

ConvNeXtV2 ~\cite{woo2023convnext} + Swin-T ~\cite{liu2021swin} 
& 28.3M & 73.92 & 74.77 & 73.87 & 74.77 & 71.75
& 95.92 & 95.97 & 95.80 & 95.97 & 91.51 \\

\textbf{ConvNeXtV2 ~\cite{woo2023convnext} + LoRA \cite{hu2022lora} (Ours)}
& \textbf{2.9M} & \textbf{85.54} & \textbf{85.05} & \textbf{84.70} & \textbf{85.05} & \textbf{83.31}
& \textbf{98.63} & \textbf{98.60} & \textbf{98.61} & \textbf{98.60} & \textbf{97.11} \\
\bottomrule
\end{tabular}%
}
\end{table*}

\Cref{tab:results} reports the quantitative comparison of the evaluated models on the GastroVision and Kvasir-Capsule datasets. We compare the proposed ConvNeXtV2 + LoRA model with several widely used convolutional architectures, including MobileNetV2, Inception V3, DenseNet121, and DenseNet201, as well as a hybrid architecture combining ConvNeXtV2 with a Swin Transformer.

On the GastroVision dataset, the proposed ConvNeXtV2 + LoRA model achieves the best performance across all evaluation metrics. Specifically, it obtains a precision of 85.54, recall of 85.05, F1-score of 84.70, accuracy of 85.05, and an MCC of 83.31. Compared with strong baselines such as DenseNet121 and DenseNet201, which achieve F1-scores of 82.12 and 81.82 respectively, the proposed method demonstrates improved classification performance. These results indicate that parameter-efficient fine-tuning using LoRA can effectively enhance the capability of the ConvNeXt backbone for gastrointestinal endoscopy image classification.

Similar trends are observed on the Kvasir-Capsule dataset. The ConvNeXtV2 + LoRA model achieves the best overall performance with a precision of 98.63, recall of 98.60, F1-score of 98.61, accuracy of 98.60, and an MCC of 97.11. While DenseNet201 also achieves strong performance with an F1-score of 98.57 and MCC of 97.07, the proposed approach slightly improves the overall results, demonstrating its effectiveness for capsule endoscopy image classification.

Overall, the experimental results show that integrating LoRA with ConvNeXtV2 consistently improves performance across both datasets, enabling efficient adaptation of large pretrained models while maintaining strong classification accuracy in challenging medical imaging tasks.

\begin{figure}[htbp]
    \centering
    \begin{subfigure}[t]{0.98\columnwidth}
        \centering
        \vskip -0.1cm
            \includegraphics[width=0.95\linewidth]{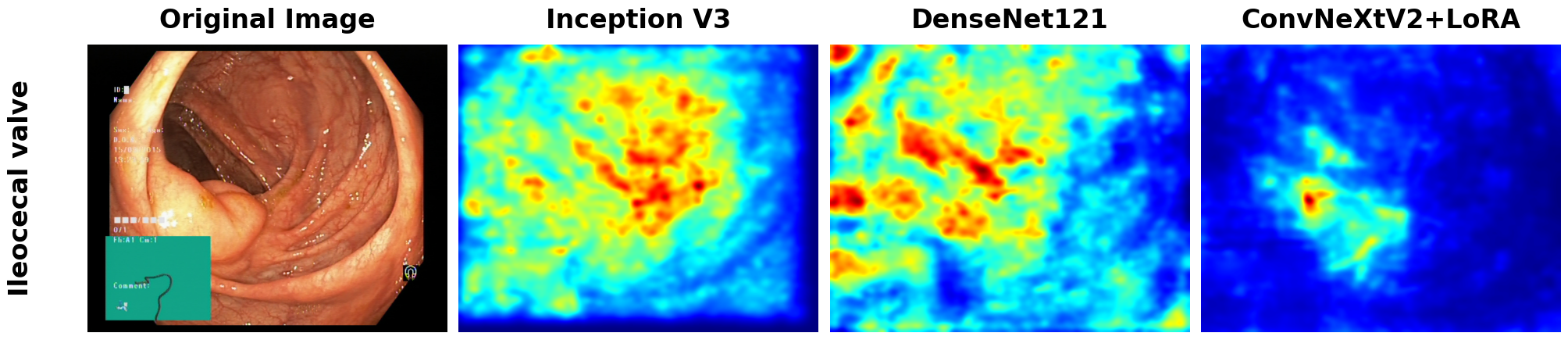}
            \includegraphics[width=0.95\linewidth]{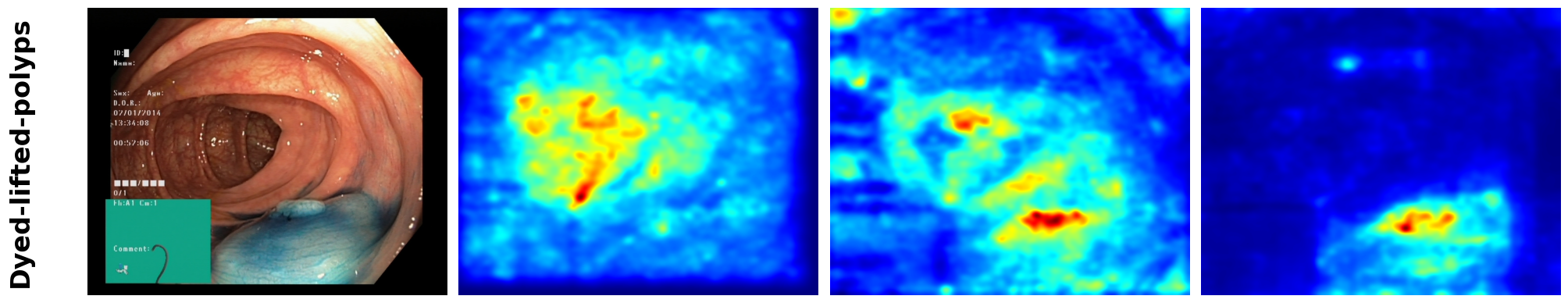}
            \includegraphics[width=0.95\linewidth]{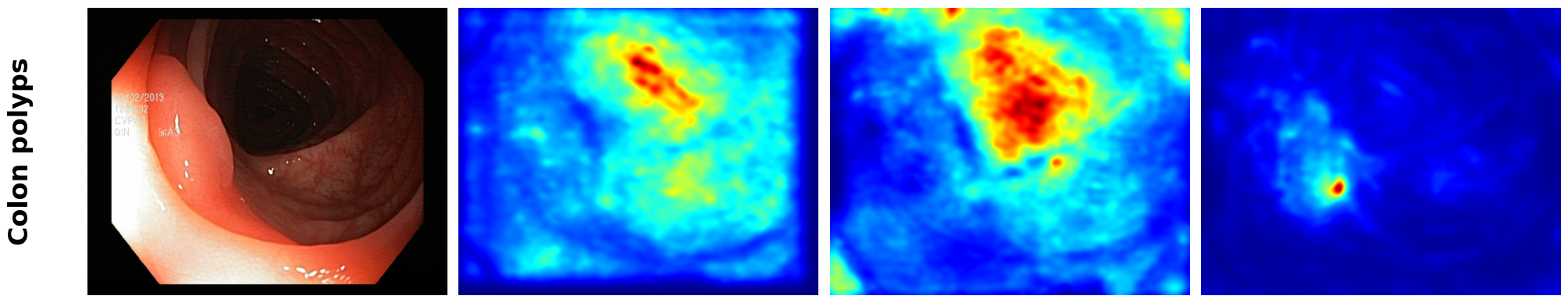}
            \includegraphics[width=0.95\linewidth]{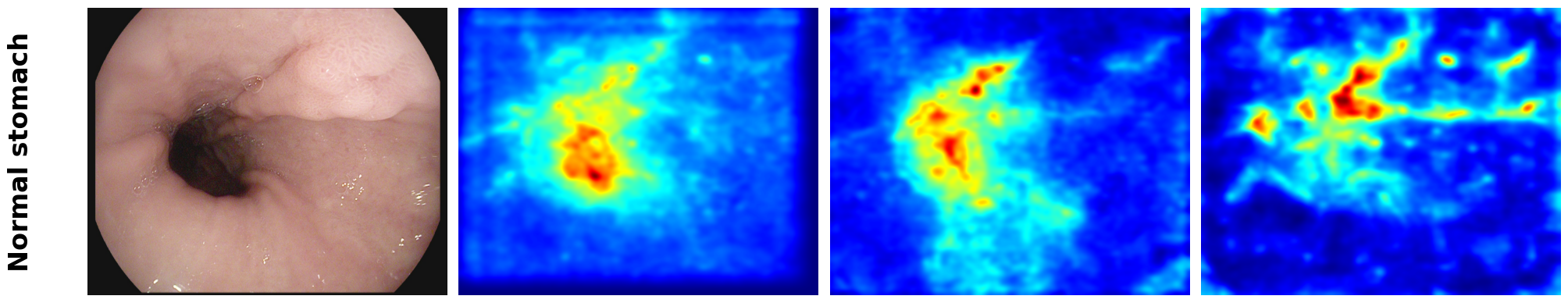}
            \includegraphics[width=0.95\linewidth]{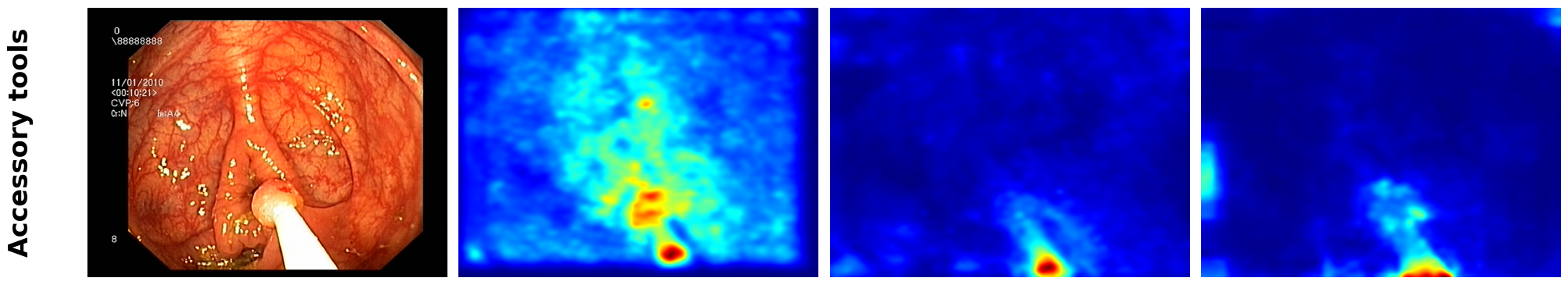}
            
        \vskip 0.03cm
    \end{subfigure}
    \vskip 0.2cm
   
    \caption{Saliency maps comparison between our method (ConvNeXtV2 + LoRA) and other baseline methods. Our method is more coherent and focuses on the semantics of the image, which a clinician uses to make the classification of the disease.}
    \label{fig:saliency_maps}
\end{figure}


\section{Conclusion}
\label{sec:conc}

 Our work's evaluation shows that our proposed approach, which combines the ConvNeXtV2 backbone with Low-Rank Adaptation (LoRA) fine-tuning, achieves better performance in classifying gastrointestinal diseases from endoscopic images. When compared to established models like MobileNetV2, Inception V3, DenseNet121, DenseNet201, and hybrid ConvNeXtV2 + Swin-T architectures, our model consistently outperformed all benchmarks. This was true across key metrics such as precision, recall, F1-score, accuracy, and MCC. On the challenging GastroVision dataset, it reached a leading accuracy of 85.05 and an F1-score of 84.70. On the Kvasir-Capsule dataset, it achieved an impressive accuracy of 98.60 and an F1-score of 98.61, confirming its robustness and reliability. These results, along with the model's efficiency and interpretability supported by saliency map visualizations (given in \cref{fig:saliency_maps}), highlight its strong potential for use in real-world clinical environments. It offers a reliable AI tool to help in the accurate and timely diagnosis of gastrointestinal conditions.

{
    \small

}


\end{document}